\pdfoutput=1

\documentclass[11pt]{article}

\usepackage[]{ACL2023}

\usepackage{times}
\usepackage{latexsym}
\usepackage{booktabs}
\usepackage{graphicx}
\usepackage{subfig}
\usepackage{multirow} 
\usepackage{CJK}
\usepackage[T1]{fontenc}
\usepackage{algorithm}
\usepackage{algorithmic}
\usepackage{comment}

\usepackage{amsmath}

\usepackage[utf8]{inputenc}
\usepackage{amsmath}
\usepackage{amssymb}
\usepackage{makecell}
\usepackage{bm}

\usepackage{microtype}

\usepackage{inconsolata}

\title{A Benchmark for Understanding and Generating Dialogue between Characters in Stories}

\author{
    Jianzhu Yao, Ziqi Liu, Jian Guan, Minlie Huang\thanks{Corresponding author}\\
    The CoAI group, Tsinghua University, Beijing, China \\
    Department of Computer Science and Technology, Tsinghua University, Beijing, China \\
    Beijing National Research Center for Information Science and Technology \\
    \{yjz19, liuzq19, j-guan19\}@mails.tsinghua.edu.cn;\\
    aihuang@tsinghua.edu.cn \\
}

\begin{document}
\maketitle
\begin{abstract}
  Many classical fairy tales, fictions, and screenplays leverage dialogue to advance story plots and establish characters. We present the first study to explore whether machines can understand and generate dialogue in stories, which require capturing traits of different characters and the relationships between them. To this end, we propose two new tasks including Masked Dialogue Generation and Dialogue Speaker Recognition, i.e., generating missing dialogue turns and predicting speakers for specified dialogue turns, respectively. We build a new dataset \textsc{DialStory}, which consists of 105k Chinese stories with a large amount of dialogue weaved into the plots to support the evaluation. We show the difficulty of the proposed tasks by testing existing models with automatic and manual evaluation on \textsc{DialStory}. Furthermore, we propose to learn explicit character representations to improve performance on these tasks. Extensive experiments and case studies show that our approach can generate more coherent and informative dialogue, and achieve higher speaker recognition accuracy than strong baselines.
\end{abstract}

\section{Introduction}
Dialogue plays an important role in various types of literary works such as short stories, novels, and screenplays by advancing plots, establishing characters, and providing expositions using natural and lifelike words~\cite{kennedy1983literature}. Compared to dialogue in conversational scenarios such as chit-chat bots~\cite{shang2015neural} or task-oriented dialogue systems~\cite{deng2012use}, dialogue in stories is mainly used to exhibit emotions, motivations, or personalities of characters following the authors' design, which further serve for the coherence, informativeness, engagingness, and plot development of whole stories. Dialogue is also revealed to be essential for user-agent interaction in many text adventure games~(\citealp{xi2021kuileixi};\citealp{li2022immersive}). It has not been widely explored for machines to understand and generate dialogue in stories despite the broad recognition of the importance of this ability. 

\begin{figure}[t]
    \centering
    \includegraphics[width=\linewidth]{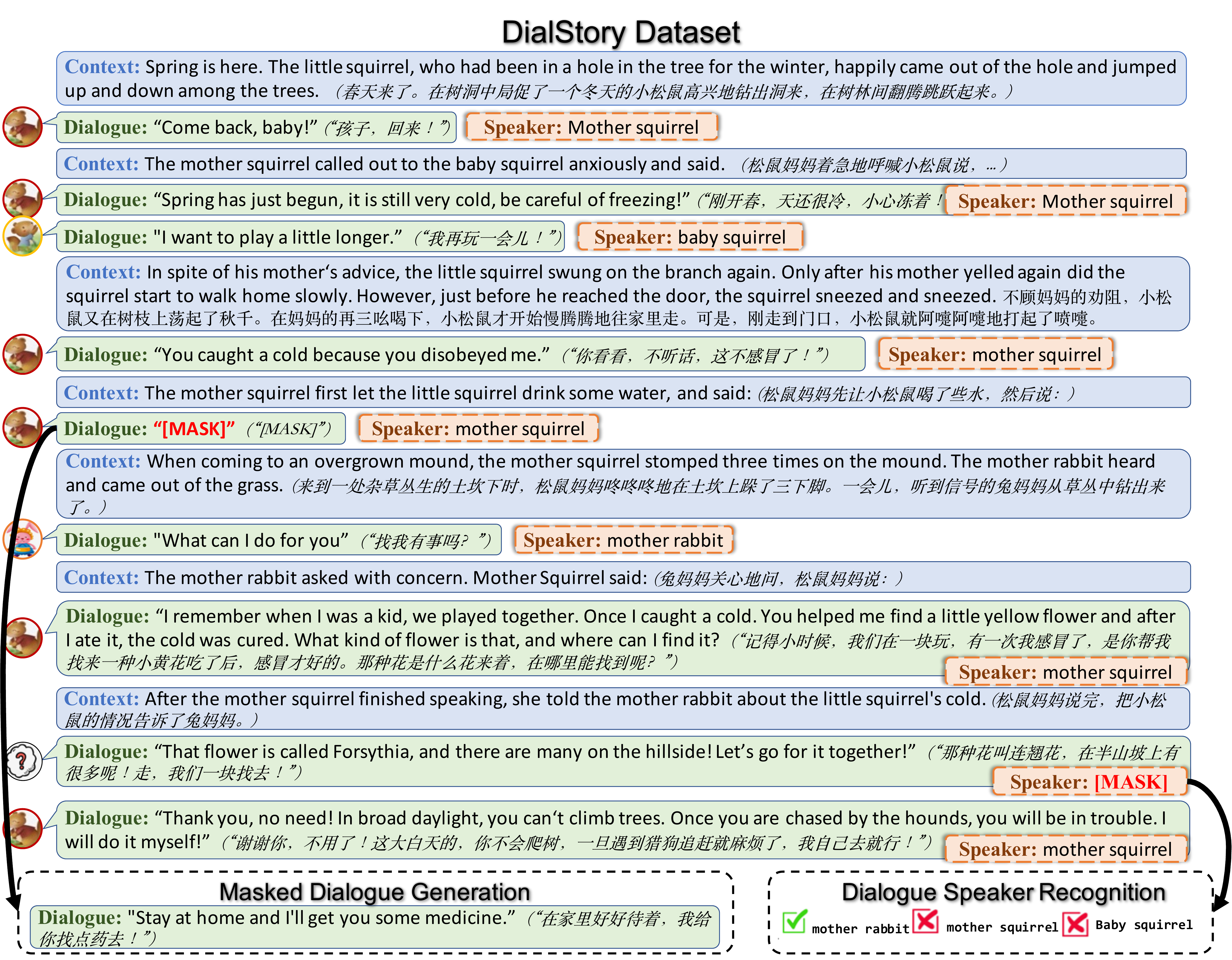}
    \caption{An example from \textsc{DialStory} and the proposed two tasks: \textit{Masked Dialogue Generation} and \textit{Dialogue Speaker Recognition}. }
    \label{data_example}
\end{figure}

To spur research in this field, we present a new story dataset named \textit{\textsc{DialStory}}, which consists of 105k Chinese short stories with automatic annotations of dialogue turns and corresponding speakers. Furthermore, we formulate two new tasks including: \textit{(1) Masked \underline{Dial}ogue \underline{Gen}eration~(DialGen):} completing a story with several dialogue turns masked with placeholders; and \textit{(2) \underline{Dial}ogue \underline{Sp}ea\underline{k}er Recognition~(DialSpk):} choosing correct speakers from given candidates for several specified dialogue turns. We construct standardized datasets for these tasks through automatic or manual annotation based on \textsc{DialStory}. As exemplified in Figure~\ref{data_example}, 
these tasks comprehensively investigate the ability to capture characters' emotions~(e.g., the \textit{mother squirrel} is worried about baby squirrel's catching cold), motivations~(e.g., the \textit{mother squirrel} intends to call her baby to come back home or get medicine), and relationship between each other~(e.g. \textit{mother rabbit} knows the kind of flower as a friend of \textit{mother squirrel} when they were kids). 

Furthermore, we found in massive stories that dialogue had a strong connection to different characters, reflected in their emotional states, speaking styles, and plot development. To provide more insights to tackle these tasks, we propose to learn character representations for modeling dependencies between characters and dialogue explicitly. Extensive experiments and case studies show that our approach can generate more coherent and informative dialogue, and achieve higher speaker recognition accuracy than strong baselines. 

\section{Related Works}

\paragraph{Story Understanding and Generation}
Recently there have been various tasks proposed to assess the ability of story understanding and generation such as story ending selection~\cite{2019Story}, story ending generation~\cite{guan2019story}, story completion~\cite{DBLP:conf/ijcai/Wang019b}, 
story character identification~\cite{brahman-etal-2021-characters-tell}, story generation from prompts~\cite{fan2018hierarchical}, titles~\cite{yao2019plan} and beginnings~\cite{guan2020knowledge}. Another line of work focused on controllable attributes in story generation such as keywords~\cite{DBLP:conf/emnlp/XuPSPFAC20}, outlines~\cite{rashkin2020plotmachines}, emotional trajectories~\cite{brahman2020modeling}, styles~\cite{kong-etal-2021-stylized} and characters' personalities~\cite{zhang2022persona}. 
Different from them, we emphasize the importance of dialogue in developing characters and maintaining the story's coherence. 
\paragraph{Dialogue in story} 
The dialogue context is comprised of a variety of messages from users and the system, determining the conversation topic and the user's goal for the conversation\citep{serban2017hierarchical}. The dialogue in the story makes this feature more apparent. 
Recently, like AI Dungeon, some text adventure games focus on players and machines co-creating stories. \citet{li2022immersive} proposed the idea that players can play the role of the chosen character and chat with others through the dialogue, while \citet{xi2021kuileixi} presented an AI game where players can explore different ways to reach the plot goals. In contrast, our tasks focus on both the understanding and generation of dialogue in the story, and they can be easily generalized to more complex AI interactive games.

\paragraph{Character Representations} Dialogue unfolds in a story around several inter-related characters. Accordingly, it is essential to capture the traits of different characters for dialogue modeling.
\citet{ji2017dynamic, clark2018neural} proposed to update characters' representations dynamically based on previous hidden outputs to track their states, which does not apply to the parallel architecture of Transformer for training. \citet{azab-etal-2019-representing} derived character representations from their corresponding dialogue turns. In contrast, we learn character representations from story plots, which then serve to understand and generate dialogue.

\section{\textsc{DialStory} Dataset}

We construct the \textsc{DialStory} dataset by randomly sampling 105k chapters from the Chinese novels released by \citet{guan2022lot} with each chapter including at least ten dialogue turns. We also set a restriction that the number of tokens in all dialogue turns should account for at least 30\% and at most 50\% of the total length of the story, in order to keep a balance between the context and dialogue. We automatically annotate dialogue turns in these stories as text spans that are surrounded by quotation marks. Then, we use a pretrained named entity recognition model~\cite{zhao2019uer} to identify all people's names. Each distinct name corresponds to a character. We also conduct a manual annotation on 150 stories, and the accuracy of character identification is 718/746=96.2\%, which shows the high quality of this automatic method. We then decide the speaker of the dialogue by recognizing the subjects of sentences before and after the dialogue turn using spaCy\footnote{\url{https://spacy.io/}}.
Table~\ref{allStory} shows the statistics of our dataset. 

\begin{table}[!ht]
    \centering
    \begin{tabular}{l|c}
    \toprule
        \multicolumn{2}{c}{DialStory}\\
        \midrule
        Avg. \#Token&451.98\\ 
        Avg. \#Dialogue Token&20.68 \\ \midrule
        Avg. \#Sentence&22.14\\
        Avg. \#Dialogue Turn&12.78\\ \midrule
        Avg. \#Character&3.54\\
        \bottomrule
    \end{tabular}
    \caption{Statistical average numbers for the \textsc{DialStory} dataset. \textit{\#Dialogue token} means the average number of tokens in each dialogue turn.}
    \label{allStory}
\end{table}
\begin{table*}[!ht]
    \centering
    \begin{tabular}{l|ccc|ccc}
    \toprule
         \multirow{2}{*}{Statistics}&\multicolumn{3}{c|}{DialGen}&\multicolumn{3}{c}{DialSpk}\\  
         &Training&Validation&Test&Training&Validation&Test \\ \midrule
         \#Examples&100k&2.5k&2.5k&20k&100&150\\
         \midrule
         Avg. \#Token~(Input)&453.64&454.12&453.77&489.51&489.25&490.49 \\ 
         Avg. \#Sentence~(Input)&21.75&21.92&21.71&28.44&28.14&28.75 \\
         Avg. \#Dialogue Turns~(input)&10.35&10.45&10.34&13.68&13.54&13.82 \\ 
         \midrule
         Avg. \#Token~(Output)&82.04&81.94&81.64&-&-&-\\ 
         Avg. \#Dialogue Turns~(Output)&3.91&3.90&3.94&-&-&-\\
         \midrule
         Avg. \#Specified Dialogue Turn&-&-&-&5.01&5.05&4.85\\
         Avg. \#Candidate Character&-&-&-&5.27&5.82&4.97\\
         \bottomrule
    \end{tabular}
    \caption{Statistics for the DialGen and DialSpk dataset.}
    \label{MDG_data}
\end{table*}

\section{Proposed Tasks}

We aim to measure model's ability to understand and generate dialogue in a story. To this end, we design the dialogue generation task Masked Dialogue Generation and dialogue understanding task Dialogue Speaker Recognition. We show the task definitions, targets, dataset construction and statistics below.

\subsection{Masked Dialogue Generation}

\paragraph{Task Formulation} 
We formulate the DialGen task as follows: given an incomplete story $S=(s_1, s_2,\cdots, s_T)$ where each $s_i$ is a token and several dialogue turns are masked with placeholder tokens $\lbrack\rm MASK\rbrack$, the model should generate the missing dialogue turns consecutively to form a coherent story. We denote the ground truth as $D=(d_1,d_2,\cdots,d_N)$ of $N$ tokens. 

\paragraph{Dataset Construction}
We use the following constraints to construct the DialGen dataset based on \textsc{DialStory}:
\begin{itemize}
    \item We randomly mask 30\% of the dialogue turns in each story.
    \item We do not mask the first 50 tokens to provide sufficient background information for the story.
    \item We do not mask the last 30 tokens to provide ending information for that story.
    \item We ensure that each input story (i.e. with masked dialogue turns) mentions at least five characters.
\end{itemize}

Table \ref{MDG_data} shows the detailed statistics.

\subsection{Dialogue Speaker Recognition}

\paragraph{Task Formulation} 
We formulate the DialSpk task as follows: given a story $S=(s_1,s_2,\cdots,s_T)$  and a set of all character names $\mathcal{C}=\{C_1,C_2,\cdots,C_K\}$ that are mentioned in $S$ as candidates, the model should choose the correct speakers from $\mathcal{C}$ for $M$ specified dialogue turns $\mathcal{D}=\{{D}_1,{D}_2,\cdots, {D}_M\}$ in $S$. 

\paragraph{Dataset Construction} We randomly sampled 20k stories from \textsc{DialStory} and automatically annotate the speaker for each dialogue turn for training, and resorted to manual annotation for validation and testing. For manual annotation, we first ask one annotator to label the characters in a story and the speaker of each dialogue turn. Then we asked another two annotators to check the correctness of the annotations, e.g., whether all mentioned characters are annotated, and whether each dialogue speaker is correct. We require the first annotator to re-annotate those examples that another two annotators do not agree on, and repeat the above process until all annotators agree on the examples. We also sampled 100 stories in the training set for manual annotation to investigate the accuracy of automatic annotation, which we will discuss in Section~\ref{dialspk-result}. Table \ref{MDG_data} shows the detailed statistics.

\section{Methodology}
\begin{figure}[ht]
    \centering
    \includegraphics[width=\linewidth]{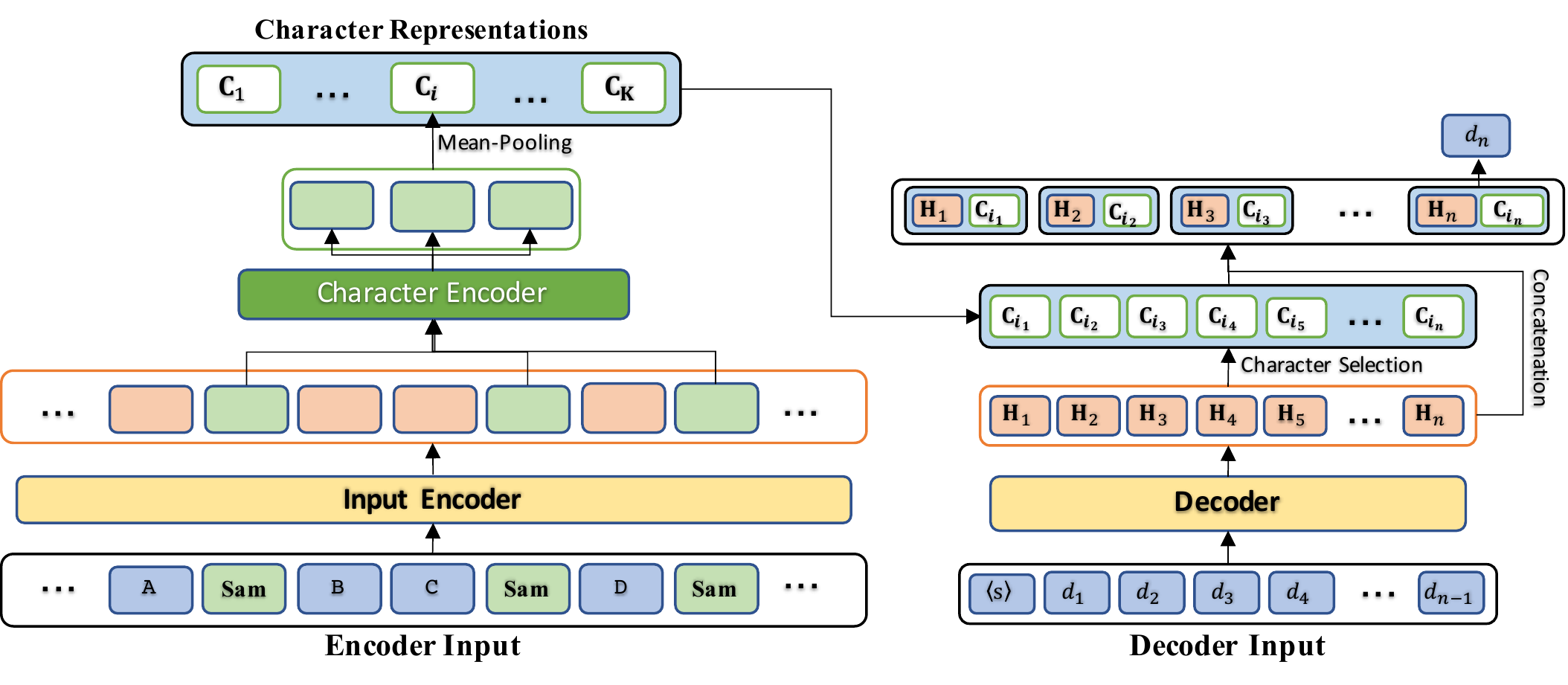}
    \caption{Model overview for the DialGen task. $\langle\rm s\rangle$ is the start-of-sequence token.}
    \label{fig:model_architecture}
\end{figure}

We propose to learn representations of different characters and exert them on decoding masked dialogue turns or predicting speakers. 
In this section, we describe the details of our model. Figure~\ref{fig:model_architecture} shows the model overview for the DialGen task.  

\subsection{Character Representation Learning}
We derive the representation of a character $C_i$, denoted as $\textbf{C}_i$, by aggregating the output hidden states of the BART encoder corresponding to all mentions of $C_i$ as follows: 
\begin{equation}
\label{align:character}
\textbf{C}_i=\text{Pool}(\text{CharEnc}(\{\textbf{h}_k|s_k~\text{mentions}~C_i\})),
\end{equation}
where CharEnc is a bi-directional Transformer character encoder, Pool means applying mean pooling on the character encoder outputs, $\textbf{h}_k$ is the hidden state of the input encoder at the $k$-th position with $s_k$ as the input token. In that way, we get representations of all the characters from the input: $\textbf{C}_1,\textbf{C}_2,\cdots,\textbf{C}_K$, where $K$ is the total number of the characters appearing in the text. And these representation vectors can be used at the decoding stage or other comprehension tasks.

\subsection{Character Representation Utilization}
\paragraph{DialGen} We incorporate learned representations of $K$ characters
into the decoder to build explicit connections between dialogue and character features. At the $n$-th time step, the decoder dynamically selects a character $C_{i_n}$ and then combines its representation $\textbf{C}_{i_n}$ for token prediction as follows:
\begin{equation}
    P(d_{n}|d_{<n})=\text{softmax}(\boldsymbol{W}f([\textbf{H}_n;\textbf{C}_{i_n}])+\boldsymbol{b}),
\end{equation}
where $\textbf{H}_n$ is the output hidden state of the decoder at the $n$-th step, $[;]$ is the concatenation operation, $\boldsymbol{W}$ and $\boldsymbol{b}$ are trainable parameters and $f$ is the SiLU activation function~\cite{elfwing2018sigmoid} followed by layer normalization. The decoder decides ${C_{i_n}}$ as follows: 
\begin{align}
\label{align-argmax}
    {i_n}&=\text{argmax}_{{i}}\hat{\textbf{H}}_n\cdot \textbf{C}_i,~~i=1,2,\cdots,K\\ 
    \hat{\textbf{H}}_n&=\boldsymbol{W}_c\textbf{H}_n+\boldsymbol{b}_c,
\end{align}
where $\textbf{C}_i$ is the representation for $C_i$, $\boldsymbol{W}_c$ and $\boldsymbol{b}_c$ are trainable parameters.
Finally, the model is optimized by minimizing the standard language modeling loss as follows:
\begin{align}
\mathcal{L}_{\rm DialGen}&=-\sum_{n=1}^N\text{log}P(d_n|d_{<n}).
\end{align}
\paragraph{DialSpk} 
\par We feed the input story $S$ into the input encoder and obtain the representations of all candidate characters following Eq.~\ref{align:character}. We use the output hidden state at the position of a special token $\lbrack\rm MASK\rbrack$, which is inserted before each specified dialogue turn $D_m$~($m=1,2,\cdots,M$), as the corresponding dialogue representation $\textbf{D}_m$. Then we optimize the following loss:
\begin{align}
    \mathcal{L}_{\rm DialSpk}&=-\sum_{m=1}^M\text{log}P(\hat{C}_{i_m}=C_{i_m}),\\
    P(\hat{C}_{i_m})&=\text{softmax}(\textbf{D}_m\cdot\{\textbf{C}_k\}_{k=1}^K),
\end{align}
where $C_{i_m}$ is the ground-truth speaker label for $D_m$~($i_m=1,2,\cdots,K$), $P(\hat{C}_{i_m})$ is the distribution of the predicted speaker overall the candidate list $\mathcal{C}$.

\begin{table*}[h]
\centering
\begin{tabular}{l|c|ccccc}
    \toprule
\textbf{Model}&\textbf{\#Param}&\textbf{BLEU1}&\textbf{BLEU2}&\textbf{DIST2}&\textbf{DIST3}&\textbf{DIST4}\\
\midrule   
BART&116M&21.04&7.85&6.67&20.57&33.78\\ 
Ours&125M&\textbf{23.88}&\textbf{8.47}&\textbf{7.72}&\textbf{25.35}&\textbf{43.10}\\ 
    \bottomrule
    \end{tabular}
    \caption{Automatic evaluation results for DialGen. \#Param is the number of parameters.}
    \label{mdg_automatic}
\end{table*}

\begin{table*}[!ht]
\small
\centering
    \begin{tabular}{c|cccc|cccc|cccc}
    \midrule
        \multirow{2}{*}{\textbf{Models}}&\multicolumn{4}{c|}{\textbf{Fluency}(\%)}&\multicolumn{4}{c|}{\textbf{Coherence}(\%)}&\multicolumn{4}{c}{\textbf{Informativeness}(\%)} \\
    
        &\textbf{Win}&\textbf{Lose}&\textbf{Tie}&$\kappa$&\textbf{Win}&\textbf{Lose}&\textbf{Tie}&$\kappa$&\textbf{Win}&\textbf{Lose}&\textbf{Tie}&$\kappa$\\
    \midrule
        \textbf{\textbf{Ours} vs. \textbf{BART}}&30.0&30.0&40.0&42.0&39.9**&15.0&45.1&42.3&41.0*&17.0&42.0&43.5 \\
    \midrule
    \end{tabular}
    \caption{Manual evaluation results. The scores indicates the percentage of \textit{win, lose}, or \textit{tie} when comparing our model with the BART baseline. $\kappa$ denotes Fleiss' Kappa \citep{fleiss1971measuring} to measure the inter-annotator agreement. * means p-value $<$ 0.05, ** means p-value $<$ 0.01 (Wilcoxon signed-rank test). }
    \label{task1-manual}
\end{table*}

\section{Experiments}
Since our approach adapts to all encoder-decoder models, we build our model based on BART$_{\rm Base}$~\cite{shao2021cpt}. We use BART$_{\rm Base}$, BERT\citep{devlin-etal-2019-bert}, RoBERTa\citep{liu2019roberta}, and MacBERT\citep{cui2020revisiting} as baselines. 
We follow BART$_{\rm Base}$'s hyper-parameters and initialize our model with the public checkpoint\footnote{\url{https://huggingface.co/fnlp/bart-base-chinese}}. Both the input encoder and decoder contain 6 hidden layers with 768-dimensional hidden states. BERT\footnote{\url{https://huggingface.co/bert-base-chinese}}, RoBERTa\footnote{\url{https://huggingface.co/hfl/chinese-roberta-wwm-ext}}, and MacBERT\footnote{\url{https://huggingface.co/hfl/chinese-macbert-base}} uses 12-layer encoder with 768-dimensional hidden states. All the models are trained on Quadro RTX 6000 GPU.

\subsection{Masked Dialogue Generation}

\paragraph{Implementation Details}
To conduct experiments on the masked dialogue generation task, we decide the hyper-parameters based on the performance of the validation set. We train \citet{shao2021cpt}'s BART model for 4.6 epochs with a 1e-4 learning rate for 1 day, and for our model, we train it for 5.6 epochs with a 1e-4 learning rate for 1 day. All baselines and our model are trained using the Adam optimizer.

During the training process for our method, we computed the selection coverage of characters within a single story. And it showed that in every 1000 training steps, the coverage of different characters ranged from 98.64\% to 99.00\%, which meant nearly all the characters are selected during training, and all the characters contributed to the generated dialogue. It further proved that the $\text{argmax}$ in Eq.~\ref{align-argmax} operation doesn't break the gradient progress when training for this task.

\paragraph{Automatic Evaluation}

Following previous works, we use several standard, widely used automatic evaluation metrics. We use \textbf{BLEU-$n$}~\cite{papineni2002bleu} to measure the average word overlap between each generated and ground-truth dialogue turn~($n$=1,2),  and \textbf{Distinct-$n$} \citep{li2015diversity} to evaluate $n$-gram diversity of generated dialogue turns~($n$=2,3,4).

We further train a classifier to evaluate the overall \textbf{coherence} by fine-tuning BERT$_{\rm Base}$~\cite{devlin2018bert} to discriminate stories whose dialogue turns are randomly reordered~(labelled with 0) from original stories~(labelled with 1)~\cite{guan2020union}. 

To be more specific, for the coherence classifier, we construct the training and validation sets by randomly shuffling the order of dialogue turns and keeping other content in the correct order. We regard the perturbed story as a negative example and the original story as a positive example. We sample another 195k stories~(except those in \textsc{DialStory}) from the novels of \citet{guan2022lot} to construct the training set~(190k examples) and the validation set~(5k examples). We train the model for 4 epochs with a 2e-5 learning rate and a 16-batch size, using the Adam optimizer. During the evaluation, we consider an example coherent when the probability of being coherent predicted by the classifier is greater than 0.5. We use the ratio of outputs~(along with the input) that are classified as coherent by the classifier to all generated outputs as the coherence score.



\par The result of the automatic evaluation is presented in Table \ref{mdg_automatic}. According to the table, compared to the BART baseline, our model consistently generates more word overlaps with ground truth and achieves better diversity under the guidance of character representations, which means our model can generate more diverse but not commonplace responses. 

\begin{figure}
    \centering
    \includegraphics[width=\linewidth]{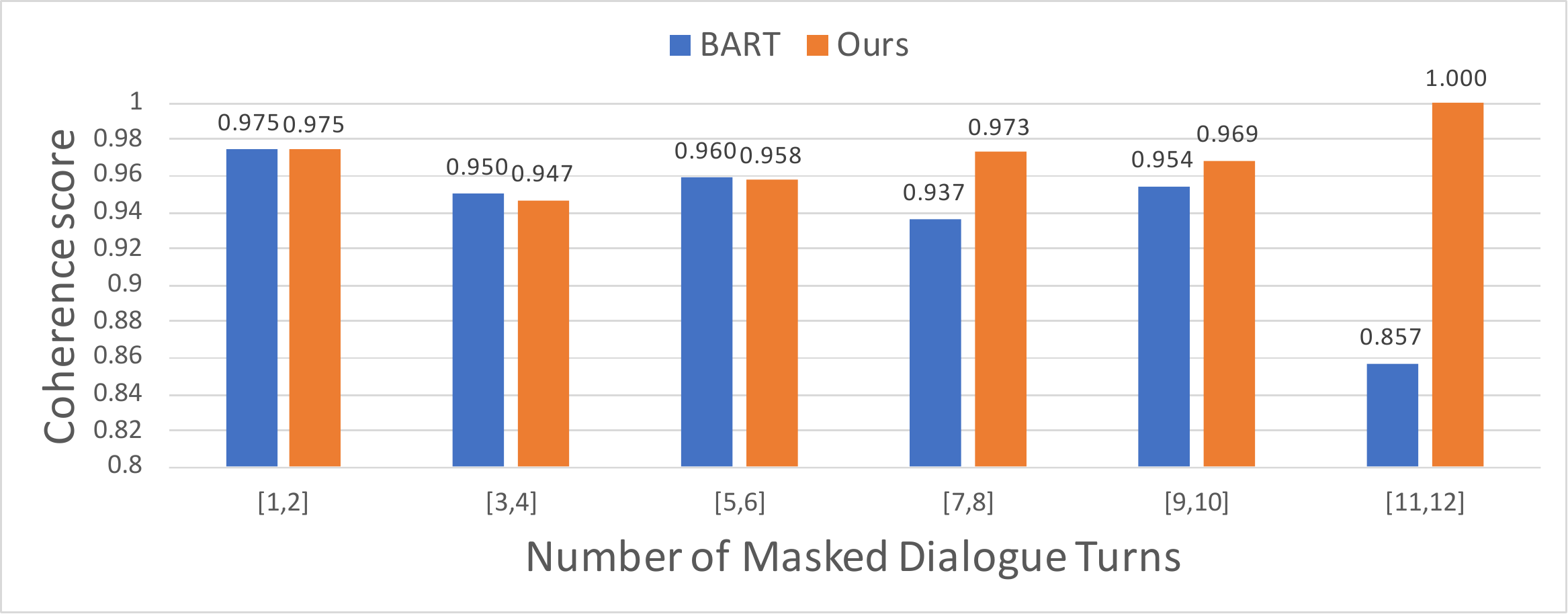}
    \caption{The coherence score of our model and BART varies with the number of masked dialogue turns.}
    \label{DiaLogic_png}
\end{figure}

Figure~\ref{DiaLogic_png} plots the coherence score varying with the number of masked dialogue turns. The result shows that our model gets a higher coherence score than BART when required to generate more than seven turns of dialogue in one story. 

\paragraph{Manual Evaluation}
We conduct a pairwise comparison between our model and the BART baseline. We randomly select 100 examples from the test set. For each pair of outputs along with the input, we ask three annotators to give a preference (\textit{win}, \textit{lose} and \textit{tie}) in terms of fluency, coherence, and informativeness. All the annotations are native Chinese speakers. We adopt majority voting to make final decisions among the annotators. The three aspects of manual evaluation are as follows:

\begin{enumerate}
    \item \textbf{Fluency:} Grammatical correctness and intra-sentence linguistic quality.
    \item \textbf{Coherence:} Inter-sentence relatedness, causal and temporal dependencies. We judge the coherence between the story and a dialogue turn by following the criterion in Table~\ref{tab:score-criterion}. We add the scores of all the generated dialogue turns in a story to get the overall coherence score of the story, which is then used to compare with each other.
    \item \textbf{Informativeness:} Interesting, diverse and rich details.
\end{enumerate}

\begin{table}[h]
    \centering
    \small
    \begin{tabular}{c|c}
    \toprule
         \textbf{Score}&\textbf{Coherence Criterion}  \\ \midrule
         0& has nothing to do with the whole story\\ \midrule
         1& with correct mentions or reference but not logical\\\midrule
         2& perfectly match the masked position \\
    \bottomrule
    \end{tabular}
    \caption{Coherence scoring criterion for each generated dialogue turn.}
    \label{tab:score-criterion}
\end{table}

As shown in Table \ref{task1-manual}, all the results show moderate ($\kappa > 0.4$) agreement, which shows our model outperforms the BART baseline significantly in dialogue informativeness and coherence.   

\begin{figure*}[!h]
    \centering
    \includegraphics[width=0.9\linewidth]{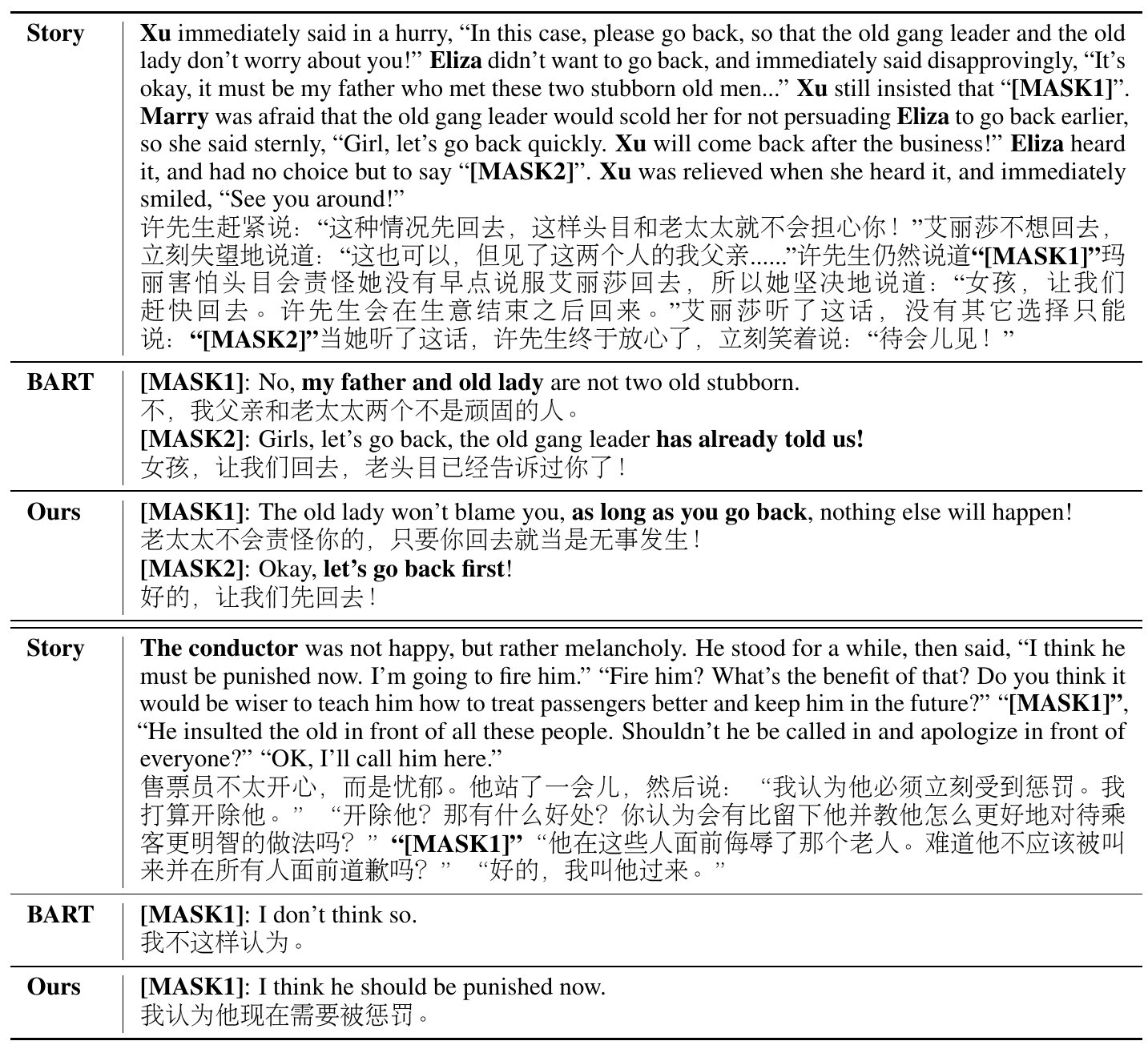}
    \caption{Case study of the DialGen task. The special tokens [MASK1], [MASK2] refer to the positions of missing dialogue turns. We highlight the different characters in bold.}
    \label{case_study_task1}
\end{figure*}

\paragraph{Case Study}
Figure~\ref{case_study_task1} showed two examples to investigate how learning character representations can help our model generate more coherent dialogue. We found that our model can better model the relationship between different characters and the direction of the storyline. For example, in the first case, we can see that the BART's generation confuses different characters' fathers, while our model captures the relationship between different characters, and generates proper responses for the corresponding characters, which also moves the plot forward. And in the second case, we can see that BART's generation is commonplace and contradicts the plot development. In contrast, our model captures the intentions of the speaker and the development trend of the plot, generating an appropriate and coherent response. Since these two models use the same pretrained weight, we can infer that the character modeling module leverages the coherent and reasonable generation.

\begin{figure}[h]
    \centering
    \includegraphics{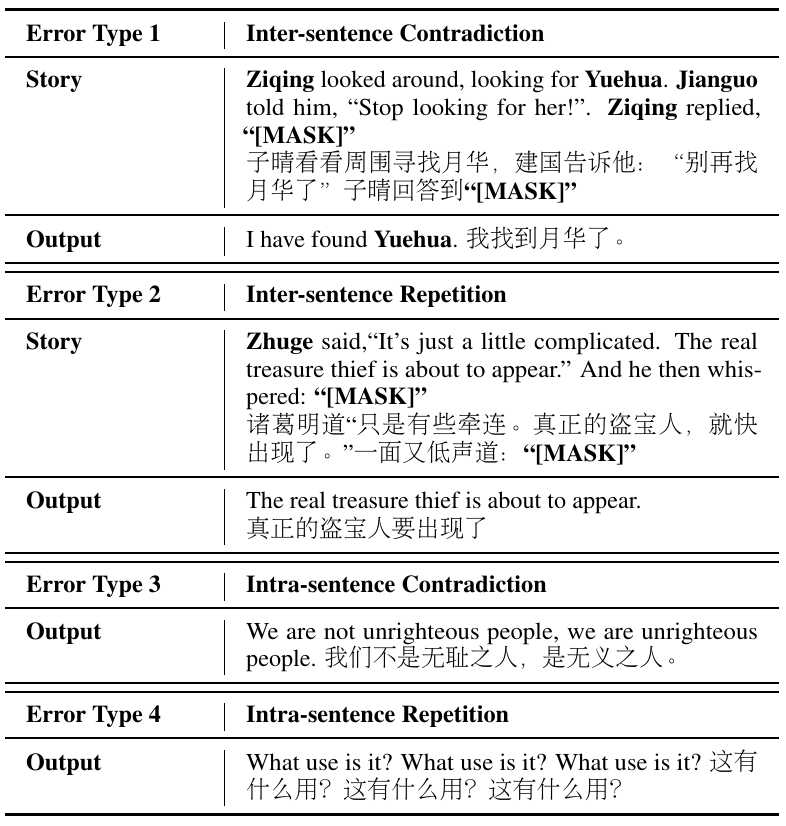}
    \caption{Cases for different error types for the DialGen task.}
    \label{dialgen_error}
\end{figure}

We also summarize four error types of the generated dialogue turn for the DialGen task: \textbf{(1)} Inter-sentence Contradiction; \textbf{(2)} Inter-sentence Repetition; \textbf{(3)} Intra-sentence Contradiction; \textbf{(4)} Intra-sentence Repetition. We show the typical corresponding cases in Figure~\ref{dialgen_error}. We conducted a quantitative analysis of those 4 error types on our model's generation. We analyzed 20 stories with 103 dialog turns and the results are shown in Figure~\ref{tab:quantitative-analysis-error}. We found that both our model and BART suffer from these errors, suggesting that there is still space for model improvement, especially in the inter-sentence repetition. 

\begin{table}[h]
\small
    \centering
    \begin{tabular}{c|c|c|c|c}
    \toprule
         \textbf{Error Type}&Type 1&Type 2&Type 3&Type 4  \\\midrule
         \textbf{Proportion}&4.85\%&16.5\%&1.94\%&2.91\% \\ \bottomrule
    \end{tabular}
    \caption{The quantitative analysis of the error types.}
    \label{tab:quantitative-analysis-error}
\end{table}

\subsection{Dialogue Speaker Recognition}
\paragraph{Implementation Details}

\par To conduct experiments on the speaker recognition task, we decide the hyper-parameters based on the performance of the validation set. For the BART baseline and our approach, we insert a mask token before each dialogue needed to be predicted, and a person id token before and after each character name span. Then, we insert all the unique person id tokens before the input stories as different options, and make predictions based on the cos similarity of option tokens and mask tokens. We train \citet{shao2021cpt}'s BART model for 30 epochs with a 5e-5 learning rate for 3 days. For encoder-only baselines, we implemented BERT, RoBERTa, and MacBERT and trained them for 15 epochs with a 1e-5 learning rate for 2 days. For our model, we train it for 22 epochs with a 1e-6 learning rate for 2 days. All baselines and our model are trained using the Adam optimizer.

\paragraph{Metrics}
We evaluate the DialSpk task using two automatic metrics including \underline{d}ialogue-level \underline{ac}curacy (DAC) and \underline{s}tory-level \underline{ac}curacy (SAC). DAC is calculated as the ratio of the correct predictions to the total number of specified dialogue turns, while SAC is the ratio of the number of stories where all dialogue turns are correctly predicted to the number of all test examples. These two metrics provide the evaluation for dialogue understanding with different granularities.

\begin{align}
\small
\textbf{DAC} = \frac{\text{\# correct predictions}}{\text{\# total number of specific dialogue turns}}\times 100\%
\end{align}
\begin{align}
\small
    \textbf{SAC} = \frac{\text{\# correct stories}}{\text{\# total number of stories}} \times 100 \%
\end{align}

\begin{table}[h]
    \centering
    \begin{tabular}{l|c|cc}
    \toprule
        \textbf{Models}&\textbf{\#Param}& \textbf{DAC(\%)}&\textbf{SAC(\%)} \\\midrule
        BERT&103M&62.00&20.00\\
        RoBERTa&103M&61.80&21.30\\
        MacBERT&103M&61.70&20.00\\
        BART&116M& 61.80&21.20\\
        \midrule
        Ours&125M&\textbf{93.30} &\textbf{74.70}\\
    \bottomrule
    \end{tabular}
    \caption{Experiment results for the DialSpk task.}
    \label{SR_re}
\end{table}

\paragraph{Results}\label{dialspk-result}
As shown in Table \ref{SR_re}, our model outperforms all the baselines significantly~($p<0.01$, Wilcoxon signed-rank test) on both DAC and SAC scores, suggesting the benefit of learning character representations. We tested the accuracy of automatic training set annotations, and the DAC/SAC scores are 86.78\%/67.80\%. Together with the model's performance on the test set, we can see the automatic annotation for the training set is of good quality. We also conducted the human prediction experiment, and the DAC/SAC scores are 97.90\%/90.70\%, which are much higher than the best model. So there is much room for further improvement for machine-based approaches.


\section{Discussion}
\paragraph{Masked Dialogue Generation} In this task, the masked positions can be anywhere in the story. To generate and complete one specific masked dialogue turn, this task requires the machine to first understand the main line of the whole story, the roles and features of different characters, and then infer and generate the most appropriate dialogue turn to move the plot forward according to the specific environment. 
Context information also plays an important role in this task, which puts forth a high demand for dialogue and plot coherence. And this task can easily leverage other research such as persona chat bots, emotional chat bots, or even some other story generation tasks.

\paragraph{Dialogue Speaker Recognition} This task specifically focuses on the dialogue understanding in a story. In a story, there are always lots of characters speaking to each other in different environments. And the relationship between characters and dialogue is complicated. Sometimes a piece of dialogue is self-talk, sometimes it is a conversation between two parties, or even multiple parties in a complex environment, and sometimes there are many details in the background information that cover up the actual speaker. The speaker of the dialogue will not only appear in the surrounding, but may also hide in the previous or behind the context of the dialogue. To recognize all the speakers in one story, our model also needs to capture different characteristics of people. Higher requirements for the model's understanding of dialogue and character relationships are also put forth by the necessity to predict speakers simultaneously in several positions. This task can also leverage other research on recognition in multi-person dialogue story scenes.

\section{Future Work}

Despite the overall improved performance of our character modeling methods, we find that there is still space for further improvement. Our approach constructs the character representations from the input stories as static vectors. But the status of characters could be updated during the generation process. As a result, we can dynamically update those representations during plot development. We plan to explore the new possibility of the character-update technique and leave it as an important future work. 

\section{Conclusion}
In this work, we present the first study on understanding and generating inter-character dialogue in stories. To this end, we collect a Chinese story dataset \textsc{DialStory} with a large amount of dialogue, and propose two new tasks including masked dialogue generation and dialogue speaker recognition. We also construct standardized datasets for these tasks through automatic and manual annotations based on \textsc{DialStory}. By incorporating representations of different characters, our model outperforms strong baselines significantly on both tasks in terms of automatic and manual evaluation. The benchmark datasets, tasks, and models will further boost the development of this field. 

\section{Limitations}

For the DialSpk task, in the test set, there are 150 stories, and a total of 728 masked dialogue positions, and in the validation, there are 100 stories and 505 positions. For the DAC score, there is enough data to evaluate the models' performance. The dataset is small to some degree for the SAC evaluation. Although the validity of our model could be reflected in this dataset, we also plan to augment this annotated dataset for future research.

\bibliography{custom}
\bibliographystyle{acl_natbib}

\appendix

\end{document}